\documentclass[conference]{IEEEtran}
\IEEEoverridecommandlockouts
\usepackage{cite}
\usepackage{amsmath,amssymb,amsfonts}
\usepackage{algorithmic}
\usepackage{graphicx}
\usepackage{textcomp}
\usepackage{stfloats}
\usepackage{siunitx}
\usepackage{xcolor}
\def\BibTeX{{\rm B\kern-.05em{\sc i\kern-.025em b}\kern-.08em
    T\kern-.1667em\lower.7ex\hbox{E}\kern-.125emX}}

\begin{document}

\title{Enabling Under-Ice Geochemical Observations with a Size, Weight, and Power-Constrained Robot\\
\thanks{J.H. was funded through the MIT UROP program, and NSF Project \#1917528. V.P. was funded through the MIT Martin Fellowship.}
}

\author{\IEEEauthorblockN{Jess Horowitz}
\IEEEauthorblockA{\textit{Mechanical Engineering} \\
\textit{Massachusetts Institute of Technology}\\
Cambridge, USA \\
jnh@mit.edu}
\and
\IEEEauthorblockN{Victoria Preston}
\IEEEauthorblockA{\textit{MIT-WHOI Joint Program} \\
\textit{Massachussetts Institute of Technology}\\
Cambridge, USA \\
vpreston@mit.edu}
\and
\IEEEauthorblockN{Anna P. M. Michel}
\IEEEauthorblockA{\textit{Applied Ocean Physics and Engineering} \\
\textit{Woods Hole Oceanographic Institution}\\
Woods Hole, USA \\
amichel@whoi.edu}
}

\maketitle

\begin{abstract}
Estimates of greenhouse gas emissions from Arctic estuarine environments are dominated by \emph{in situ} summer-time ice-free dissolved gas measurements due to the logistical ease of performing field observations in these conditions. Recent evidence in coastal Arctic environments, however, has demonstrated that dissolved methane (CH$_4$) and carbon dioxide (CO$_2$) are strongly seasonally variable, and at least one significant gas ventilation event occurs during the spring freshet. Whether the Arctic serves as a source or sink of greenhouse emissions has significant implications on modeling climate change and its feedback mechanisms. To enable higher resolution spatiotemporal measurements of dissolved gases in typically undersampled conditions, remotely operated vehicles (ROVs) can be used to extract near continuous water samples below ice before and during the spring freshet. Here, we present a size, weight, and power-constrained (SWAP) underwater vehicle (UV) and novel geochemical sampling system suitable for taking under-ice geochemical observations and demonstrate the proposed system in a field-analog setting for Arctic estuarine studies. 
\end{abstract}


\section{Introduction}
The natural environment is the largest general contributor of greenhouse gasses to the atmosphere through geochemical and biochemical processes (e.g., decay, hydrate dissolution, fauna activity) \cite{delmotte2021ipcc}. Anthropogenic sources of methane (CH$_4$) and carbon dioxide (CO$_2$) have notably disrupted natural source-and-sink cycles, accelerating the effects of climate change \cite{delmotte2021ipcc}. The Arctic marine coastal continuum (connecting inland lakes to the open ocean by networks of groundwater, rivers, and estuaries) is currently experiencing rapid changes in annual sea ice cycles, retreat of permanently frozen ground, and lengthening of the ice free season as a direct result of positive feedbacks of climate change \cite{manning2020river, magnuson2000historical, stroeve2012arctic, zona2016cold}. However, the majority of gas distribution studies performed in Arctic waters are performed in summer-time ice-free conditions e.g., \cite{fenwick2017methane}, missing important ephemeral events such as the spring freshet, which has been observed to drive over 95\% of all greenhouse gas emissions in some environments \cite{manning2020river}.

\begin{figure}[t]
    \centering
    \includegraphics[trim={0 0 0 70cm}, clip, width=1\columnwidth]{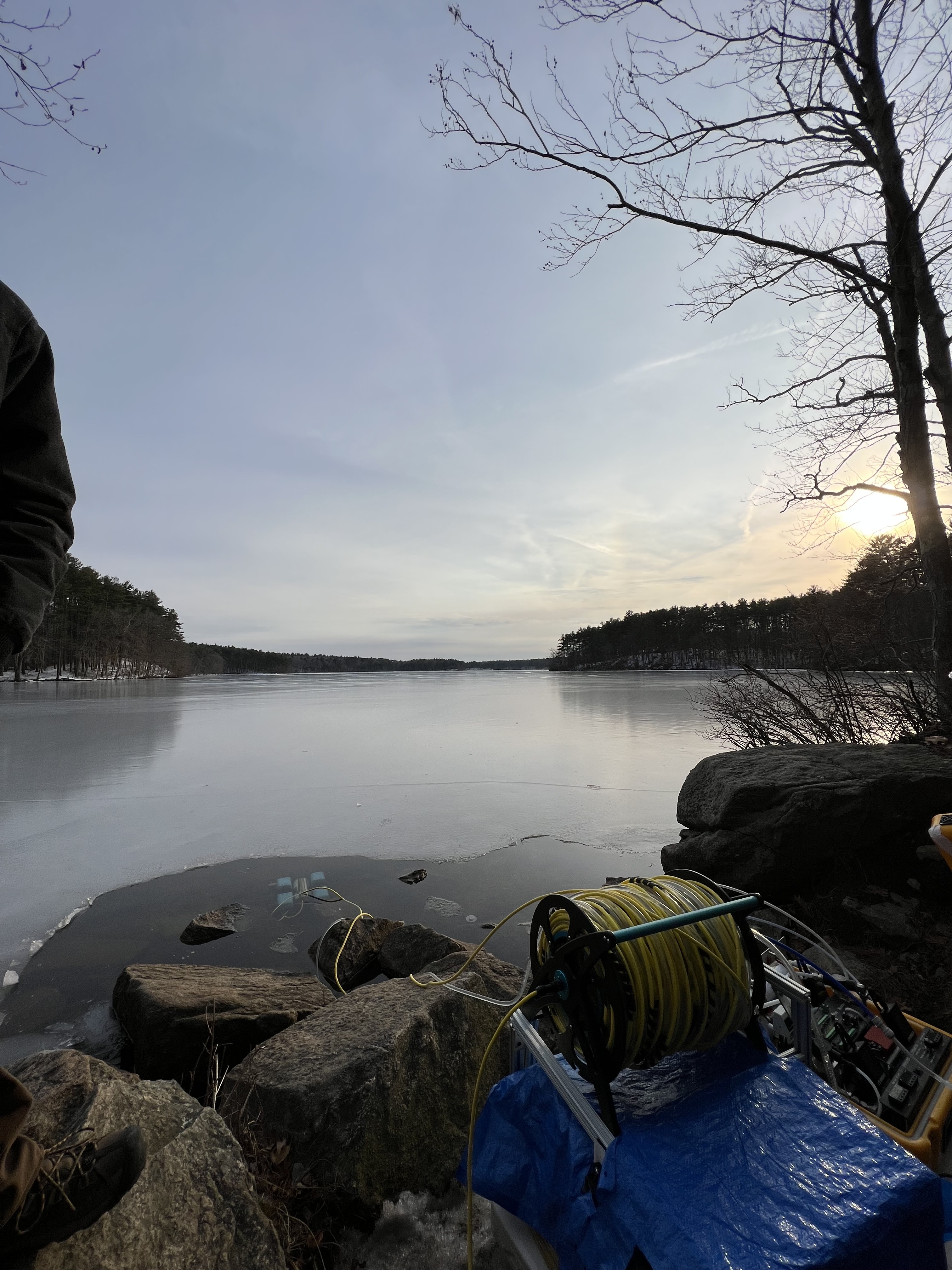}
    \caption{An image of the SWAP-UV, topside geochemical laboratory, and the ITMSS deployed at a frozen lake (Lake Whitehall, MA, USA) during an Arctic-Analog field trial.}
    \label{fig:image}
\end{figure}

Robotic technologies are uniquely well-suited to improve the spatiotemporal resolution of dissolved gas studies in these environments over short-lived events. Equipped with \emph{in situ} sensors suitable for measuring dissolved gasses and other relevant geochemical quantities (e.g., temperature, salinity), remotely operated and autonomous robots can be used to collect many tens of thousands of observations over several days. Previous studies in the Arctic have used a surface vehicle \cite{manning2020river} to investigate surface distributions of CH$_4$ and CO$_2$ in open-waters during the start of the ice-free season. The restriction to ice-free waters using this type of sampling strategy misses the true beginning of the spring freshet, in which river waters begin to flow underneath an ice cap, and restricts the vehicle to tracking up to but not underneath a receding ice edge.

To complete the picture of a spring freshet outgassing event with respect to riverine transport, we propose using a size, weight, and power-constrained (SWAP) underwater vehicle (UV) coupled to a topside gas extraction and analysis instrument. Unlike ship-operated remotely operated vehicles (ROVs) typical in marine work, SWAP-UVs can be deployed by a single operator by hand, require little logistical overhead, are typically self-powering (i.e., with a battery pack), and can carry only a few kilograms of weight. For Arctic estuarine environments, in which the gas-saturated layer has been observed to be only a few meters deep (trending higher concentrations in fresher waters of the halocline) \cite{manning2020river,bussmann2013distribution}, the small package of a SWAP-UV is advantageous for taking useful observations within the upper water column, in addition to easy through-ice deployment. The significant challenge of using a SWAP-UV then is carrying \emph{in situ} instruments for dissolved gas analysis, the most precise being both too large and too heavy for a SWAP-UV due to the use of spectroscopic equipment, active gas extraction (e.g., pumps), and large external power resources (e.g., car lead acid battery). To overcome this, we designed a novel topside pumping mechanism---the Integrated Tether Management Sampling System (ITMSS)---which allows these geochemical instruments to remain at the surface while the SWAP-UV carries an inlet tube along the communications tether for sample collection. This pumping mechanism can also be used to draw water samples for bottling and subsequent laboratory \emph{ex situ} analysis.

\begin{figure*}[t!]
    \centering
    \includegraphics[width=0.85\textwidth]{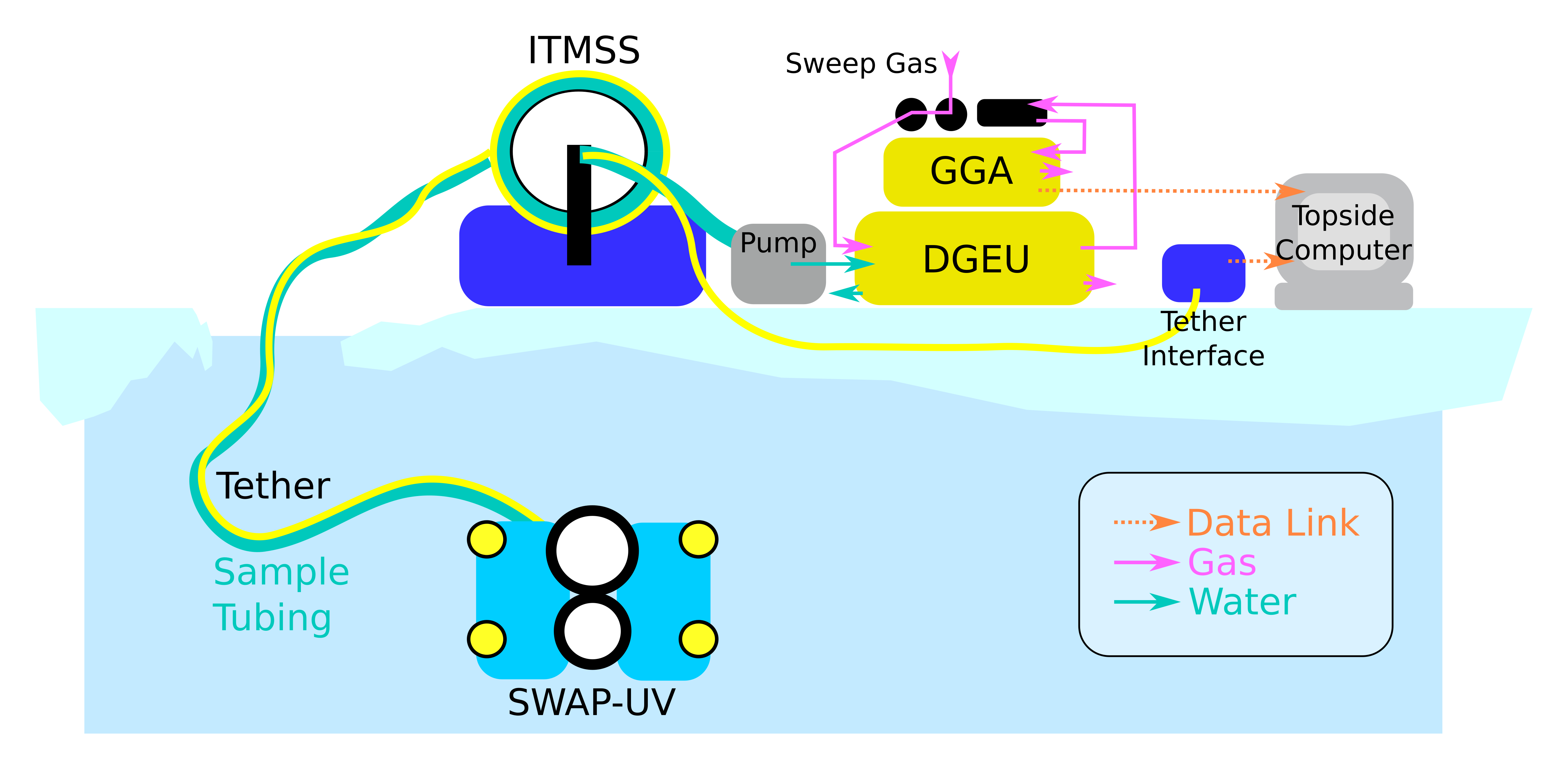}
    \caption{A conceptual illustration of the SWAP-UV, ITMSS, and topside geochemical laboratory and their connections during a field deployment for under-ice geochemical observation collection. In practice, the ITMSS and geochemical laboratory would be mounted to the back of ATVs, an ice-sled, or on the floor of a watercraft while in operation. The SWAP-UV can be deployed through a carved hole in ice; however, functionally during the spring freshet, natural holes/crevices in the ice, near-shore melting, or open waters are available as opportunistic places to deploy.}
    \label{fig:concept}
\end{figure*}

Here, we present the design of our SWAP-UV and ITMSS for gas extraction and analysis to enable under-ice geochemical measurements, and demonstrate the system in an Arctic-analog environment (a frozen pond in Massachusetts, USA) in advance of a research campaign in Summer 2022 to Cambridge Bay, Nunavut, Canada (Fig.~\ref{fig:image} and Fig.~\ref{fig:concept}). We conclude with a brief discussion of the use of SWAP-UVs for geochemical observations and suggest areas for future work.

\section{Related Works}
This study is a continuation of work completed in the Summer of 2018 in Cambridge Bay, Nunavut, Canada to study methane levels in an estuary continuum during the spring freshet \cite{manning2020river}. In this study, a remotely operated surface vehicle called a ChemYak \cite{nicholson2018rapid} was used to take \emph{in situ} observations of salinity, temperature, dissolved methane, dissolved carbon dioxide, and dissolved oxygen in the ice-free regions of a river feeding into the bay over the course of several days. By the nature of the ChemYak as a surface vehicle, ice-free conditions are operationally necessary, and so geochemical observations at the time were taken up to the rapidly receding ice edge spatially, and temporally were focused on the days following the recession of ice from the river banks. Thus, the period of time before the ice began receding, and geochemical transport under the ice while the ice was receding, was not examined, and is the subject of interest for our study.

The use of ROVs for biogeochemical examination of bays and lakes has been well-established (e.g., \cite{inskeep2015geomicrobiology}), including in under-ice or subglacial contexts (e.g., \cite{vogel2008subglacial, katlein2017new, jakuba2018teleoperation}). Many of these vehicles are relatively large, requiring a support vessel for deployment and would not be classified as ``SWAP'' robots. In contrast, most SWAP-UVs for examination of lakes and rivers have largely served in a reconnaissance capacity, such as visual sensors that are used to complement other forms of \emph{in situ} sampling (e.g., \cite{block2021vertical}). Some SWAP-UVs have been designed and demonstrated for under-ice studies (e.g., \cite{teece2009inexpensive, bolsenga1989rov}), but generally carried small chemistry probes that were not capable of measuring dissolved greenhouse gases like CH$_4$ and CO$_2$, if chemistry probes were carried at all. To this end, our proposed system has been developed to specifically extend the utility of SWAP-UVs in under-ice contexts in Arctic estuaries as geochemical sensors. 

\section{Methodology}
Our under-ice geochemical sampler consists of three key components -- a SWAP-UV, the ITMSS, and a topside geochemical laboratory. The SWAP-UV is used to take continuous samples moving laterally under an ice sheet and to perform precisely controlled vertical mapping of the stratified water column. The ITMSS, composed of a water pump and motorized winch, manages the communication tether and draws water from the inlet attached to the SWAP-UV. The geochemical laboratory, composed of a gas extraction unit and gas analyzer, performs the \emph{in situ} gas analysis of interest. A conceptual diagram of all the pieces is presented in Fig.~\ref{fig:concept}. Operationally, the system can be mounted on the back of all-terrain vehicles (ATVs) or onto an ice-suitable sled for from-shore or on-ice deployments; similarly the components easily fit on small watercraft and the SWAP-UV can be deployed over the side of the craft while in open waters.

\subsection{SWAP-UV}
Our SWAP-UV core platform is a standard BlueROV 2.0 from BlueRobotics\footnote{bluerobotics.com}. An integrated temperature sensor, pressure sensor, compass, and camera system (complemented with external lights) are the primary default sensors available on the vehicle for this study, although previous studies have shown how the BlueROV platform can be extended to carry other small package sensors (e.g., \cite{menapace2021long, girdhar2019streaming}). 

\subsection{Topside Geochemical Laboratory}
The topside geochemical laboratory consists of a Los Gatos Research dissolved gas extraction unit (DGEU), and Los Gatos Research greenhouse gas analysis instrument (GGA)\footnote{lgrinc.com}. These instruments are equivalent to the instruments used in the ChemYak autonomous surface vehicle used in the prior Arctic estuarine study \cite{manning2020river, nicholson2018rapid}, and details about these instruments can be found in the cited papers.

\subsection{Integrated Tether Management Sampling System}
\begin{figure}[t!]
    \centering
    \includegraphics[width=1\columnwidth]{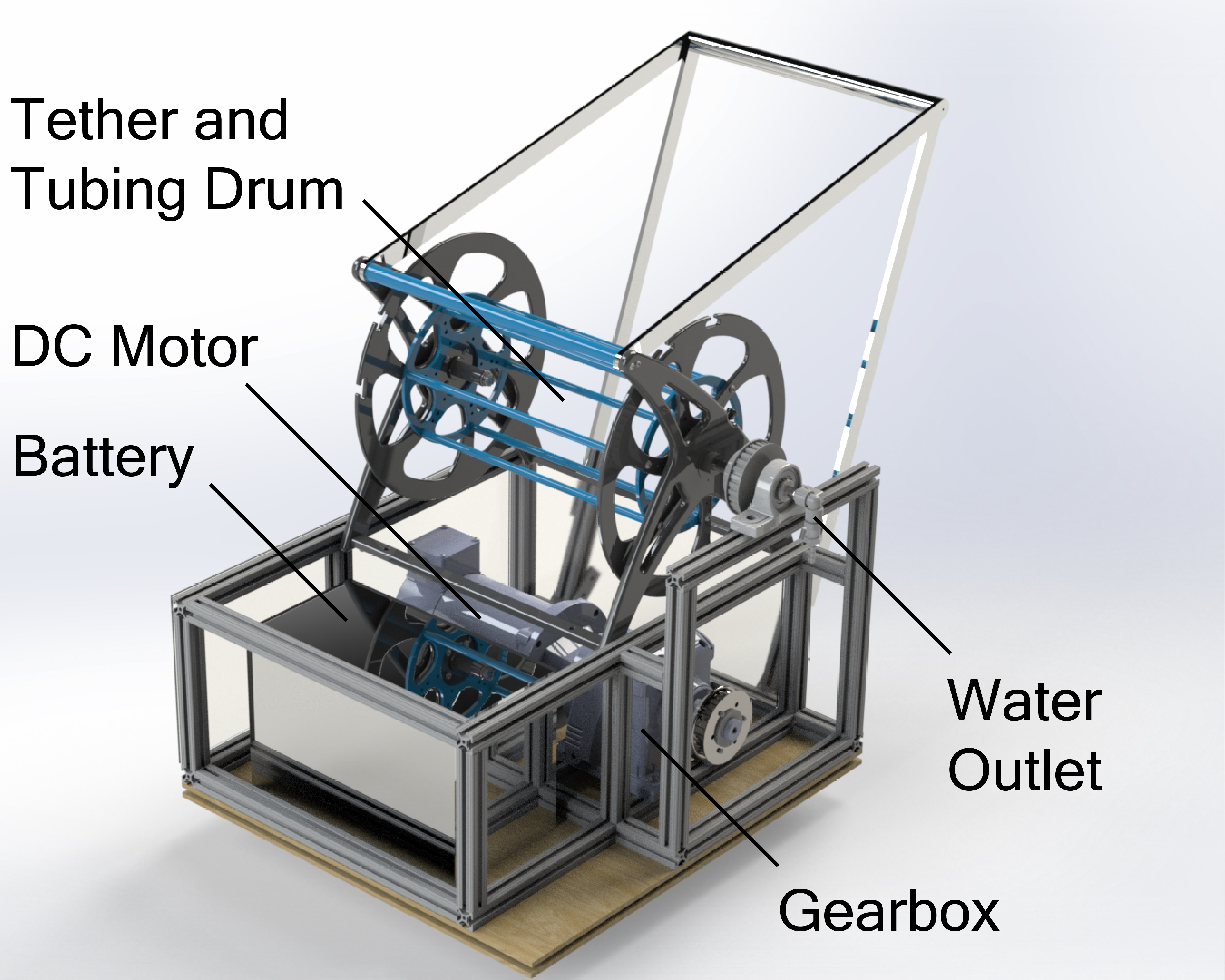}
    \caption{Full ITMSS design rendering, without tether or tubing. The frame of the ITMSS is 80/20 T-Slot Extrusion aluminum, and the motor, gearbox, and battery are mounted at the bottom of the frame to act as a counterweight when the SWAP-UV is pulled or lifted. Electronic communication passes through the left of the drum and water passes through the right. The pulley belt and water pass-through subsystem on the right side is supported by a bearing. The entire assembly is mounted on top of a rotating platform which can be pinned in place or left to freely rotate during operations.}
    \label{fig:diagram}
\end{figure}

To utilize the topside geochemical instruments, tubing must be run between the DGEU and the SWAP-UV. The ITMSS (Fig.~\ref{fig:diagram}) is designed to streamline operations and manage the SWAP-UV communications tether and DGEU tubing during deployment. Modularity is a key constraint; the ITMSS must offer portability, ease of maintenance, and adaptability to various types of field operations. It must have the ability to be securely mounted to a small vessel or deployed on an ice-sled. It must remain mechanically sound while in use and durable over multiple uses. Environmental factors necessitate that the ITMSS withstands temperatures as low as \SI{-15}{\celsius} and is both waterproof and corrosion resistant. For our intended field work, the SWAP-UV will operate in the top \SI{5}{\meter} of the water column and horizontally cover a radius of at least \SI{120}{\meter} from the deployment point. Therefore, conservatively the drum of the ITMSS should support up to \SI{150}{\meter} of coupled tether and tubing. The SWAP-UV must be retrieved every 1-2 hours for recharging. The winch must reel the vehicle at a speed of up to \SI{1}{\meter\per\second} for rapid retrieval and have a battery capacity that allows for up to 3 hours of active operation.

The ITMSS consists of a Blue Robotics Fathom Spool (hereafter ``spool'') affixed to an 80/20 T-Slot Extrusion aluminum frame for modularity. The spool is driven by a \SI{372.85}{\watt} (0.5 Horsepower) \SI{188.5}{\radian\per\second} (1800 revolutions per minute) permanent magnet DC motor with a 10:1 gearbox and powered on-board by a \SI{12}{\volt} \SI{110}{\ampere\hour} SLA battery. The communication tether for the SWAP-UV and soft PVC inlet tube (an inert polymer that will not interfere with the chemical measurements) are coupled and wound around the drum of the spool. A slip ring and fluid rotary union allow both electronic communication and sampled water to respectively pass through the spool to prevent interference with winch rotation. A \SI{12}{\volt} \SI{5}{\ampere} \SI{0.13}{\liter\per\second} (2 gallons per minute) diaphragm pump pulls water from the inlet of the PVC tube attached to the SWAP-UV up to the surface, around and through the spool, and into the DGEU of the topside geochemical laboratory. The system features an anodized aluminum pulley arm to guide the coupled tether and the arm, spool, motor, and power are all mounted to a turntable to rotate and track the SWAP-UV. The pulley arm latches onto the existing handle of the spool, and is both retractable and height adjustable for improved modularity.

The maximum load the winch must overcome is the combined friction of the spool and drag of the neutrally buoyant SWAP-UV through water in the case that it must be rapidly recovered. To compute this, we assume a linear, additive relationship between these sources of friction,

\begin{equation}
    F_D = \frac{1}{2}C_D A \rho V^2
\end{equation}
\begin{equation}
    \tau_o = \tau_f + R F_D
\end{equation}

\noindent where $C_D$ is the coefficient of drag (1.05), $A$ is the surface area (\SI{0.086}{\meter\squared}), $\rho$ is the density of water (\SI{997}{\kilo\gram\per\meter\cubed}), $V$ is water velocity (\SI{1}{\meter\per\second}), $F_D$ is the friction due to drag (\SI{45}{\newton}), $\tau_f$ is the frictional torque (\SI{0.762}{\newton\meter}), $R$ is the maximum spool radius (\SI{0.168}{\meter}), and $\tau_o$ is the desired operating torque (\SI{8.32}{\newton\meter}).

The motor specification was ultimately selected based on a desired linear reeling speed of \SI{1}{\meter\per\second} through water, and therefore, a motor operating speed of \SI{10.5}{\radian\per\second} at the expected load was computed by,

\begin{equation}
    \Omega_o = \frac{V}{r}
\end{equation}
\begin{equation}
    P_\gamma = 4\Omega_o\tau_o
\end{equation}

\noindent where $\Omega_o$ is the desired rotational speed, $r$ is the minimum spool radius (\SI{0.0952}{\meter}), and $P_\gamma$ is the rated motor power. At a 10:1 gear ratio, the \SI{373}{\watt} motor would operate at \SI{9.42}{\radian\per\second} at maximum power, which is satisfactory. The winch was tested lifting the dead weight of the SWAP-UV out of water, which would be beneficial for lifting the vehicle onto a research vessel or through an ice hole in the field. The ITMSS was able to lift \SI{4.11}{\kilo\gram} at 27\% power capacity at a comfortable speed. The motor drives the spool via a pulley belt of length \SI{0.9}{\meter} or 94 pitches.

\begin{figure}[t!]
    \centering
    \includegraphics[width=1\columnwidth]{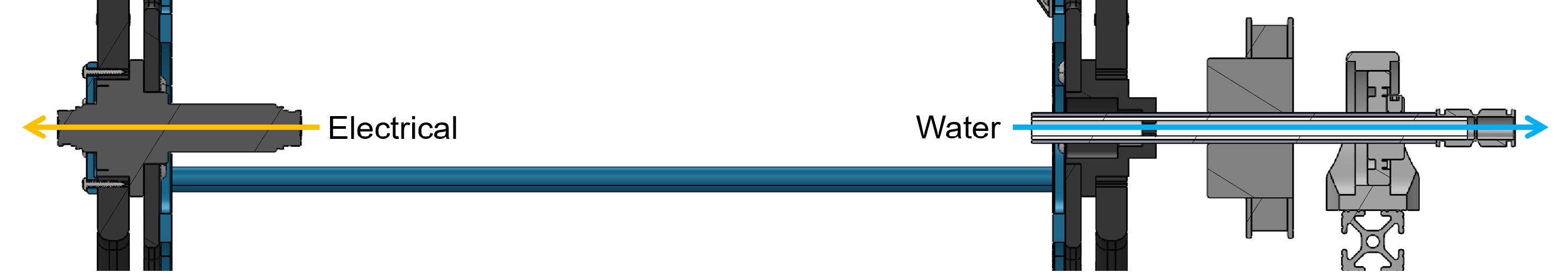}
    \caption{Cross section of ITMSS. Tethered electrical communications flow out of the drum through a slip ring. On the opposite side of the drum, water flows through LLDPE tubing within a hollow driven shaft.}
    \label{fig:crosssection}
\end{figure}

The specification of the pump must support a head of 3-\SI{5}{\meter} and losses of up to \SI{150}{\meter} of spooled \SI{0.635}{\centi\meter} inner diameter soft PVC tubing. The selected diaphragm pump in-line with the DGEU pump produced flow at the desired velocity. To allow for water to pass through the spool drum, a hollow steel shaft was mounted on its rotation axis as shown in Fig.~\ref{fig:crosssection}. Since the belt pulley is mounted at the same location as the water pass-through, a hollow steel shaft with LLDPE tubing inserted through is fastened by a shaft collar mounted to a \SI{0.24}{\centi\meter} thick steel sheet mounted to the rotating drum. To avoid cantilever bending, the shaft is supported with a corrosion-resistant acetal ball bearing mounted to the frame.

Materials for the ITMSS were selected to allow for corrosion resistance, temperature resistance, strength, and waterproofing. Specifically:
\begin{itemize}
    \item A belt was selected over a chain/sprocket assembly for driving the reel as a belt is more corrosion resistant and withstands lower temperatures without needing lubricant.
    \item The ITMSS is covered in a fitted water-shedding polyethylene tarp and light-weight corrugated plastic.
    \item The retractable pulley arm is anodized aluminum.
    \item A wireless humidity monitor is used in a sensitive electronics areas.
\end{itemize}

The ITMSS is powered by a lead-acid battery which is positioned as a counterweight to the torque of potential loads on the pulley arm, which supports the SWAP-UV tether. The spool is positioned above the motor and gearbox to make the system compact, since transportability of this system is key for field operations. This also effectively raises the overhang of the pulley arm, preventing potential drag of the tether across the surface of the ice or against a boat or sled hull. The weight is distributed such that a turntable could be mounted at the center of mass located close to the base of the ITMSS.

\begin{figure}[t!]
    \centering
    \includegraphics[width=1\columnwidth]{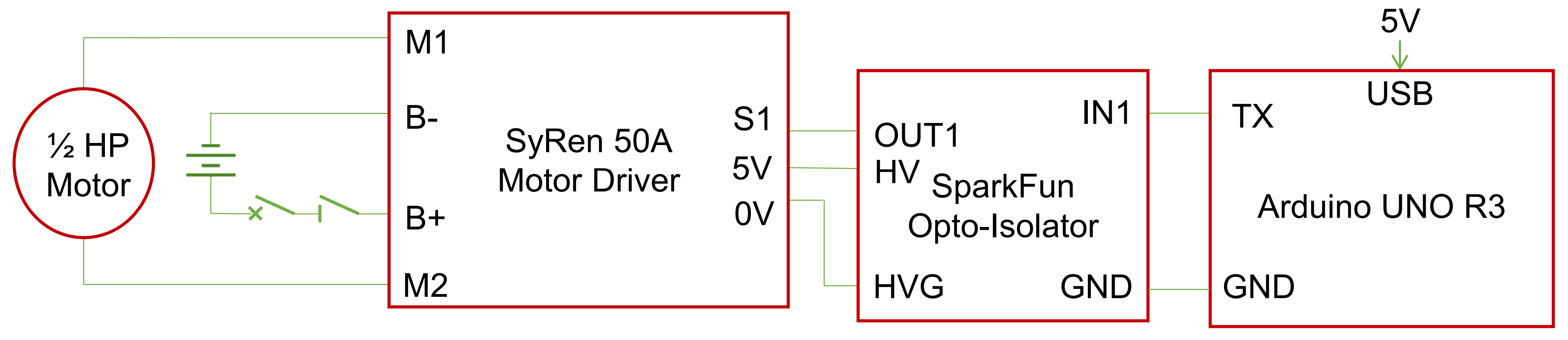}
    \caption{Electrical schematic diagram of ITMSS.}
    \label{fig:electronics}
\end{figure}

The ITMSS is controlled via an Arduino Nano which communicates with a SyRen Motor Driver using the SyRen/Sabertooth Serial library. A Python graphical user interface (GUI) allows the user to set the power and direction virtually using a slider. The Arduino script ramps up or ramps down the motor based on the GUI-specified input (represented as an integer value) corresponding to motor power. As a safety feature, a maximum power value is hard-coded into the Arduino script, preventing the specified input value from exceeding it. 22 AWG wire connects the Nano, an opto-isolator, and motor driver; 10 AWG wire supports the maximum current draw of \SI{39}{\ampere} from the \SI{12}{\volt} DC motor. A circuit breaker and an emergency stop button are integrated for safety, and mounted in an accessible location near the motor underneath the spool. The electrical schematic is shown in Fig.~\ref{fig:electronics}.

\section{Arctic-Analog Field Demonstration}
The SWAP-UV and topside geochemical laboratory tethered by the ITMSS were deployed in March 2022 in Lake Whitehall, Massachusetts, USA. Due to several weeks of cold, winter temperatures, the lake was frozen, with an ice thickness of \SI{0.05}{\meter} at the shore, and an estimated 0.1-\SI{0.15}{\meter} at the far extent of the field site. Lake Whitehall is a historic reservoir approximately \SI{2.5}{\kilo\meter\squared} with several inputs from surrounding wetlands and a cedar swamp. Most of the Lake is approximately \SI{1.8}{\meter} deep.\footnote{mass.gov/doc/whitehall-reservoir} Unlike target Arctic estuarine sites (e.g., Cambridge Bay, Nunavut, Canada), this location experiences intra-seasonal thawing, and does not have a significant surface current. This means that the test site has a unique gas distribution character in comparison to the target Arctic sites, although the physical characteristics (e.g., on and under ice) are sufficient for testing the capabilities of the SWAP-UV, ITMSS, and laboratory. 

\begin{figure*}[h!]
    \centering
    \includegraphics[width=1.0\linewidth]{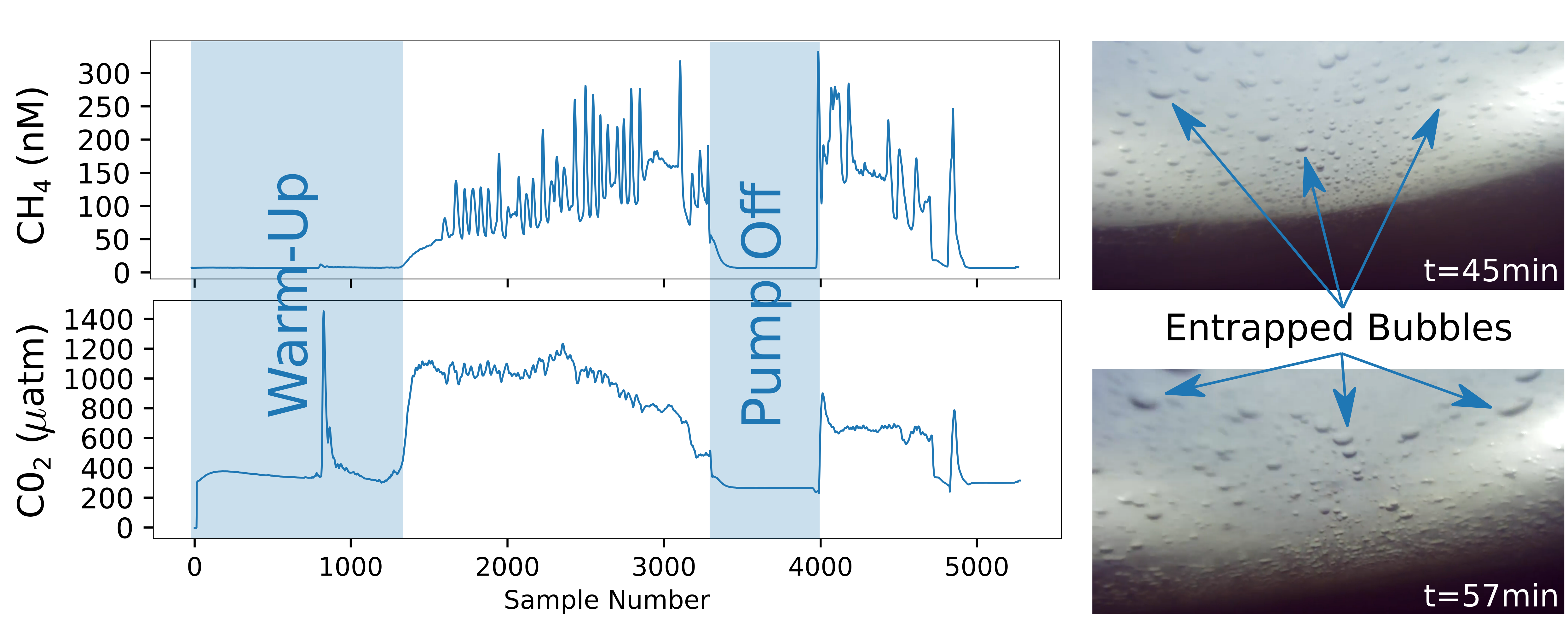}
    \caption{Methane and carbon dioxide measurements observed during Arctic analog trials. A period of instrument warm-up on shore with periodic water pumping to test the system, and a time in which the DGEU pump was turned-off during the transects when too much sediment was in the line, are indicated as light-blue regions; all other regions are valid observational periods. A peak of 300 nM CH$_4$ and 1250 $\mu$atm CO$_2$ were observed. Two still images taken from video recorded by the SWAP-UV show bubbles trapped in the ice sheet as the robot explored.}
    \label{fig:results}
\end{figure*}

\begin{figure}[t!]
    \centering
    \includegraphics[width=0.9\columnwidth]{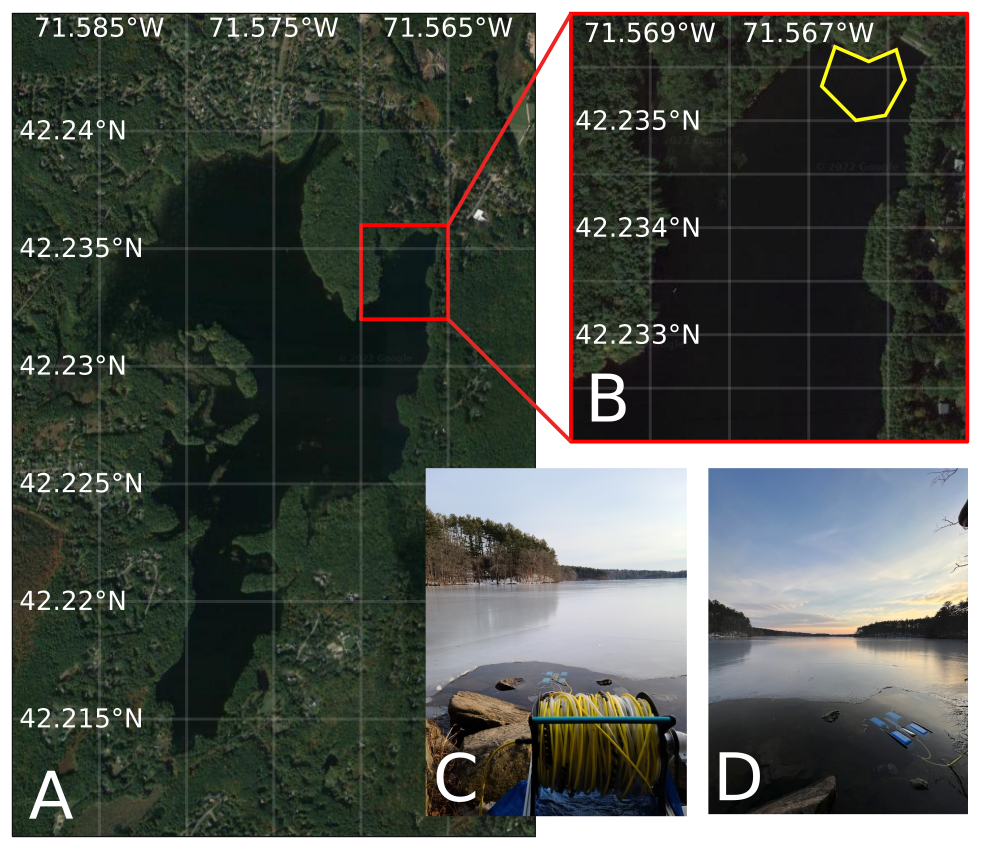}
    \caption{Satellite imagery (provided by Google Tiles in the Cartopy library) of Lake Whitehall (A and B), with estimated sample region enclosed by the yellow polygon (B). Representative pictures of lake conditions during the field deployment are shown (C and D).}
    \label{fig:map}
\end{figure}

The ITMSS and laboratory were deployed on the shore, at the location indicated in Fig.~\ref{fig:map}. A hole in the ice was made at the shore that was large enough for the SWAP-UV to be fully-submerged and propel itself away from the shore bottom. A RBR CTD was deployed at this location to measure background water temperature and salinity (\SI{2.0}{\celsius} and 0.00 PSU respectively). Using the SWAP-UV camera, the vehicle was remotely operated to swim near the ice ``ceiling'' underwater, and directed in large arcs. As the SWAP-UV has no means to exactly localize underwater, an estimated sampling region is drawn in Fig.~\ref{fig:map} based on empirical visual contact with the vehicle under the ice. The ITMSS was operated by a second remote driver, which monitored the tether and set the winch speed to let out or pull in slack.

The SWAP-UV was operated for approximately 1 hour before the battery in the vehicle was fully depleted. In this time, the laboratory took 3500 measurements of CH$_4$ and CO$_2$ from under the ice. Some measurements were missed due to sediment being sucked into the sampling line; the laboratory was decoupled from the sampling line at the surface until the water ran clear, and then replumbed directly (this was approximately 700 measurements, or 12 minutes of total sampling time). In the laboratory measurements, we see significantly elevated methane and carbon dioxide compared to atmospheric levels (Fig.~\ref{fig:results}); with a peak observation of approximately 300 nM CH$_4$, and 1250 $\mu$atm CO$_2$ (expected background levels are approximately 4 nM and 400$\mu$atm, respectively). We additionally see a rise and fall of absolute methane concentration during the trajectory corresponding with sending the robot out under the ice, staying under the ice, and returning to shore. We hypothesize the trend may be indicative of ventilation at the shores where ice was thinner and thawing in weekly cycles, in contrast to more steadfast ice at the center of the lake potentially keeping methane trapped underneath. Qualitatively, we observed bubbles in and trapped under the ice using the SWAP-UV camera, which we additionally hypothesize may be disassociated methane and other gasses.

\section{Conclusions and Future Work}
Here, we presented a modular system for taking near real-time gas observations under ice with a SWAP-UV. The ITMSS integrated tether and tubing management system was designed to support pumping water to the surface for analysis by field-deployable dissolved gas extraction unit and greenhouse gas analyzer. We demonstrated the system in an Arctic analog scenario, showing that the system successfully can be deployed in realistic outdoor conditions, can pump a sufficient stream of water for gas analysis, and measure interesting gas structures under ice, in addition to filming qualitative ice conditions. Recently (Summer 2022), a version of this system was deployed in the Arctic for under-ice measurements and exploration of melt ponds. Future work will focus on the interpretation of these data, in addition to improving SWAP-UV capabilities (e.g., localization, building more sophisticated sensors) and ITMSS adaptability for the field (e.g., swapping tethers, reducing overall weight). Fundamentally, SWAP-UVs and creative support infrastructure for maximizing their sensing abilities, show considerable promise for Arctic and general estuarine biogeochemical exploration. As the Arctic rapidly changes due to accelerating climate impacts, collecting spatially and temporally rich datasets of greenhouse gas ventilation in order to determine how these gases are produced, where they are transported or consumed, and the scale of these processes will be critical.    

\section*{Acknowledgment}
The authors would like to recognize the assistance of B. Colson, S. Youngs, W. Pardis, and other members of the Chemical Sensors Laboratory at Woods Hole Oceanographic Institution for their assistance with field campaigns and system development. We would also like to thank D. Fan and H. Schmidt at the MIT tow-tanks for the use of their tow-tank facilities for preliminary testing of the SWAP-UV. 

\bibliography{main.bib}{}
\bibliographystyle{IEEEtran}

\end{document}